\documentclass[letterpaper,10pt,journal,twoside]{IEEEtran}
\usepackage{amsmath,amsfonts}
\usepackage{algorithmic}
\usepackage{algorithm}
\usepackage{array}
\usepackage{tabularx}
\usepackage{booktabs}
\usepackage{multirow}
\usepackage[caption=false]{subfig}
\usepackage{textcomp}
\usepackage{stfloats}
\usepackage{xcolor}
\usepackage{xspace}
\usepackage{url}
\usepackage{verbatim}
\usepackage{graphicx}
\usepackage[normalem]{ulem}
\usepackage{cite}
\newcommand{\method}{AutoPath\xspace}

\title{AutoPath: Learning Transferable Goal-Conditioned Stochastic Path Prior for Safe Navigation Without Human Demonstrations}

\author{Ziyang Zhang$^{1}$, Boyang Zhou$^{1}$, Zesong Yang$^{1}$, Haocheng Peng$^{1}$, Zeming Gai$^{2}$, Xiao Liang$^{1}$, \\ Yujun Shen$^{3}$, Danping Zou$^{4}$, Ruizhen Hu$^{5}$, Hujun Bao$^{1}$, and Zhaopeng Cui$^{1\dag}$% <-this % stops a space
\thanks{Manuscript received: March 10, 2026; Accepted May 6, 2026.
This paper was recommended for publication by Editor Aniket Bera upon evaluation of the Associate Editor and Reviewers' comments. This work was partially supported by the NSFC (No.~62572425) and Ant Group.}
\thanks{$^{\dag}$Corresponding author: Zhaopeng Cui.}
\thanks{$^{1}$Zhejiang University.
        {\tt \{zhangtzeyang, byzhou, zesongyang0, haochengpeng, 22521125, baohujun, zhpcui\}@zju.edu.cn}}%
\thanks{$^{2}$Harbin Institute of Technology.
        {\tt 2023311f24@stu.hit.edu.cn}}%
\thanks{$^{3}$Ant Group.
        {\tt shenyujun0302@gmail.com}}%
\thanks{$^{4}$Shanghai Jiao Tong University.
        {\tt dpzou@sjtu.edu.cn}}%
\thanks{$^{5}$Shenzhen University.
        {\tt ruizhen.hu@gmail.com}}%
\thanks{Digital Object Identifier (DOI): see top of this page.}%
}

\begin{document}

\maketitle

\definecolor{mydarkblue}{rgb}{0,0.08,1}
\definecolor{mydarkgreen}{rgb}{0.02,0.6,0.02}
\definecolor{mydarkred}{rgb}{0.8,0.02,0.02}
\definecolor{mydarkorange}{rgb}{0.40,0.2,0.02}
\definecolor{mypurple}{RGB}{111,0,255}
\definecolor{myred}{rgb}{1.0,0.0,0.0}
\definecolor{mygold}{rgb}{0.75,0.6,0.12}
\definecolor{mydarkgray}{rgb}{0.66, 0.66, 0.66}
\definecolor{Brown2}{RGB}{238,59,59}
\definecolor{url_color}{RGB}{42, 83, 163}
\definecolor{myBlack}{rgb}{0.0, 0.0, 0.0}

\newcommand{\hc}[1]{\textcolor{mygold}{#1}}
\newcommand{\haocheng}[1]{#1}

\newcommand{\before}[1]{\iffalse#1\fi}
\newcommand{\zzy}[1]{#1}

%%%%%%%%%%%%%%%%%%%%%%%%%%%%%%%%%%%%%%%%%%%%%%%%%%%%%%%%%%%%%%%%%%%%%%%%%%%%%%%%
\begin{abstract}

Real-time navigation in cluttered and dynamic environments requires collision-free and dynamically feasible motion under limited perception. However, feasible navigation behaviors are inherently multimodal because multiple paths may exist around obstacles. In this paper, we formulate navigation as learning a transferable goal-conditioned stochastic path prior that models a reusable distribution over goal-aligned geometry-consistent local paths conditioned on local observations. This formulation enables structured sampling of navigation candidates, allowing multiple feasible paths to be explored through sampling without relying on robot-specific motion constraints. To this end, we introduce a goal-aligned canonical state representation that removes in-plane rotational ambiguity and normalizes local geometry with respect to the goal, enabling rotation-invariant path distribution learning. We further develop a structured prior learning framework that parameterizes local paths using a geometry-aware polar action manifold and incorporates risk-sensitive utility shaping with multi-goal distributional rollouts for stable and safety-aware planning. Extensive experiments in dense static environments and dynamic pedestrian scenarios demonstrate that the proposed method achieves consistently high success rates with competitive efficiency while enabling cross-platform transfer of a single path prior learned on differential-drive robots to quadruped platforms without retraining. % \looseness=-5 

\end{abstract}

\begin{IEEEkeywords}
Motion and Path Planning, Integrated Planning and Learning, Collision Avoidance.
\end{IEEEkeywords}

%%%%%%%%%%%%%%%%%%%%%%%%%%%%%%%%%%%%%%%%%%%%%%%%%%%%%%%%%%%%%%%%%%%%%%%%%%%%%%%%

\section{INTRODUCTION}

\IEEEPARstart{R}{eal-time} navigation requires generating collision-free and dynamically feasible trajectories under limited perception and continuously changing environments. In cluttered scenes, feasible motion is inherently multimodal: multiple paths may exist around obstacles, and safe navigation depends on structured spatial reasoning rather than a single deterministic action. Classical model-based planners offer explicit constraint handling but rely on hand-crafted heuristics or computationally intensive search. Learning-based approaches improve adaptability by predicting controls or trajectories directly from observations~\cite{drl_low1,drl_low2,drl_low3,crowdsurfer,pathrl}; however, many existing methods collapse this multimodality into deterministic outputs and tightly couple spatial reasoning with platform-specific motion constraints. 
Such entanglement limits reuse across robot embodiments and hinders transferable geometric reasoning across platforms.

% \IEEEpubidadjcol

\begin{figure}[!t]    
\centering        
\vspace{0.0 em}
\includegraphics[width=0.9\linewidth]{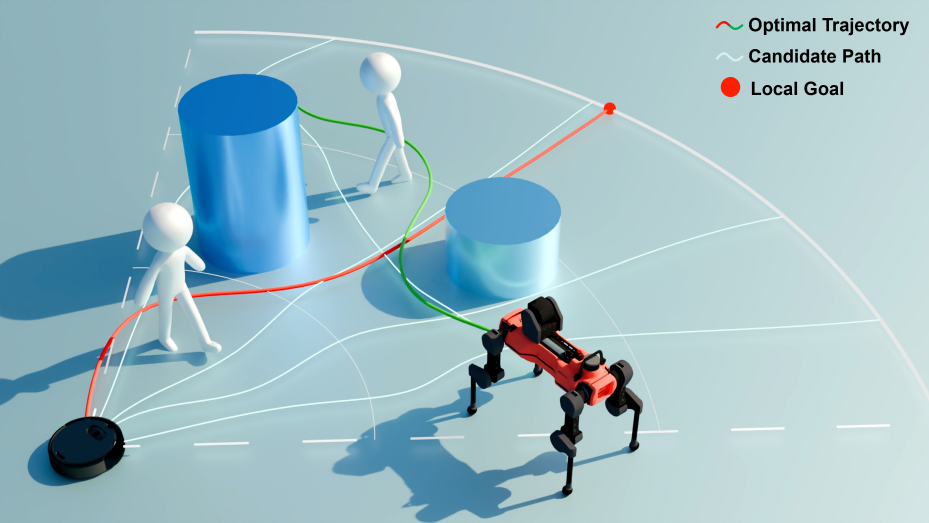}   
\vspace{-0.5 em}
\caption{
A transferable goal-conditioned stochastic local-path prior models geometry-consistent local path distributions without robot-specific motion constraints. During inference, circle-based multi-goal sampling and distributional rollouts generate diverse navigation candidates, which are refined via constrained trajectory optimization for safety and feasibility.
}
\label{fig:intro} 
\vspace{-1.5 em}
\end{figure}

% \begingroup
% \looseness=-1
In this work, we formulate navigation as learning a \emph{transferable goal-conditioned stochastic path prior}. Instead of predicting low-level controls or a single trajectory, we model a reusable distribution over geometry-consistent local paths conditioned on the goal direction and local perception. As shown in Fig.~\ref{fig:intro}, the proposed formulation operates in path-distribution space, enabling structured sampling of diverse feasible paths and supporting multimodal planning through multi-goal conditioned sampling. This perspective shifts navigation from deterministic trajectory prediction to distribution-level geometric reasoning, providing a unified foundation for adaptable and reusable planning. % \looseness=-1
% \par
% \endgroup

To enable transfer across robot platforms, we propose a transferable geometric path prior built on a new goal-aligned canonical state representation that models geometry-consistent local paths independent of robot kinematics. Specifically, we align the local observation with the goal direction to construct a canonical state frame, removing in-plane rotational ambiguity and normalizing local geometry with respect to the goal. This canonicalization transforms navigation into a rotation-invariant distribution learning problem and enables flexible provisional goal selection at test time, improving generalization across platforms and unseen environments. 
Moreover, the canonical state representation enables supervision from physics-informed planner-generated trajectories during prior learning. By expressing local observations in a goal-aligned canonical frame, these trajectories become orientation-invariant in the canonical frame, 
allowing them to serve as supervisory signals for prior learning and eliminating the need for human demonstrations. % \looseness=-1

To learn the transferable geometric path prior in a stable and expressive manner, we further design a novel structured prior learning framework. Specifically, we parameterize local paths using a geometry-aware polar action manifold with radially ordered constraints, embedding geometric structure directly into the path space to reduce the effective search dimension and stabilize learning in high-dimensional continuous control. We also incorporate risk-sensitive utility shaping to encourage safety-aware exploration during training. At inference time, the learned stochastic prior enables structured multimodal planning through circle-based multi-goal sampling and distributional rollouts conditioned on provisional goals.

Extensive experiments in dense static and dynamic environments demonstrate that AutoPath achieves high success rates with competitive efficiency and empirical transfer of the learned path prior \zzy{across tested robot platforms.} Furthermore, a single learned prior transfers from differential-drive robots to quadruped platforms without retraining, validating the effectiveness of the transferable path prior formulation.

The main contributions of this work are:
\begin{itemize}
    \item We propose \method, a novel navigation method that learns a transferable goal-conditioned stochastic path prior, enabling multimodal path generation and reusable geometric reasoning across robot embodiments.
    
    \item We introduce a goal-aligned canonical state representation that removes in-plane rotational ambiguity and normalizes local geometry with respect to the goal direction, improving generalization across platforms and unseen environments.
   
    \item We develop a structured prior learning framework that parameterizes local paths using a geometry-aware polar action manifold and incorporates risk-sensitive utility shaping with multi-goal distributional rollouts, enabling stable prior learning and robust multimodal planning.
    
    \item Extensive experiments in dense static and dynamic environments demonstrate that \method achieves consistently high success rates with competitive efficiency and cross-platform transfer of the learned path prior.
    
\end{itemize}

\section{RELATED WORK}

%\subsection{Model-based Methods}
\noindent \textbf{Model-based Methods.} 
Model-based navigation typically includes path search, local planning, and trajectory generation and optimization. Path search uses graph-based methods such as D* Lite~\cite{classical_Dlite} and JPS~\cite{classical_JPS}, or sampling-based methods such as PRM~\cite{classical_PRM} and RRT~\cite{classical_RRT}, to obtain feasible paths. Local planners such as APF~\cite{classical_APF} and DWA~\cite{classical_DWA} support real-time obstacle avoidance. Trajectory generation and optimization incorporate constraints to produce collision-free, executable, trackable trajectories~\cite{classical_tra1, classical_tra2, classical_tra3}. Despite strong stability and controllability, these methods typically require frequent parameter tuning for different tasks or environments, which limits their adaptability in dynamic or complex scenarios. In contrast, our method learns a transferable goal-conditioned stochastic local-path prior, reducing the reliance on explicit modeling and manual parameter tuning, and enabling effective transfer across different robotic platforms and environments.

% \subsection{Learning-based Methods}

\begingroup
\looseness=-1
\noindent \textbf{Learning-based Methods.}
With the rapid growth of deep learning, imitation learning (IL) and deep reinforcement learning (DRL) have been widely used for robot navigation. IL learns navigation strategies from expert demonstrations and can be deployed quickly in new scenes, but it relies on large labeled datasets~\cite{IL1, IL2} that are expensive to collect and annotate~\cite{IL4}, and it often generalizes poorly under distribution shift~\cite{IL3}.
DRL learns policies through interactive training, offering greater flexibility and adaptability, but it suffers from costly exploration~\cite{DRL_cost} and unstable control caused by policy stochasticity and environmental uncertainty~\cite{drl_low4}. To simplify control, some studies predict high-level paths instead of low-level actions~\cite{trajectory, DRL_path}, but the resulting expansion of the action space lowers sample efficiency. Axial partitioning constraints~\cite{pathrl} partly alleviate this issue but reduce adaptability in complex spatial layouts. Overall, learning-based navigation still faces major challenges in generalization, training cost, and policy stability. To address them, we formulate navigation as learning a transferable goal-conditioned stochastic path prior that models geometry-consistent local path distributions, improving generalization and robustness across diverse environments. 
\par
\endgroup

% \subsection{Hybrid Methods}
\noindent \textbf{Hybrid Methods.}
To combine the strengths of both paradigms, recent work has explored hybrid approaches. Several studies integrate learning-based strategies with Model Predictive Control (MPC) to improve stability and safety~\cite{go-mpc, ac-mpc}, but MPC still depends on accurate system models, which limits applicability~\cite{ac-mpc}. Some methods therefore shift from direct model-based control to path optimization. For example, CrowdSurfer~\cite{crowdsurfer} generates reference trajectories with learning-based methods and refines them by optimization to balance performance and computation. However, it still relies on teleoperated demonstrations during learning, which narrows the training distribution and limits generalization. In contrast, our method learns a transferable goal-conditioned stochastic path prior from physics-informed planner-generated trajectories, which avoid extensive human demonstrations.
%During inference, circle-based multi-goal sampling and distributional rollouts generate diverse navigation candidates, which are further refined through constrained trajectory optimization. \looseness=-2

\begin{figure*}[!ht]
\centering
\includegraphics[width=\textwidth]{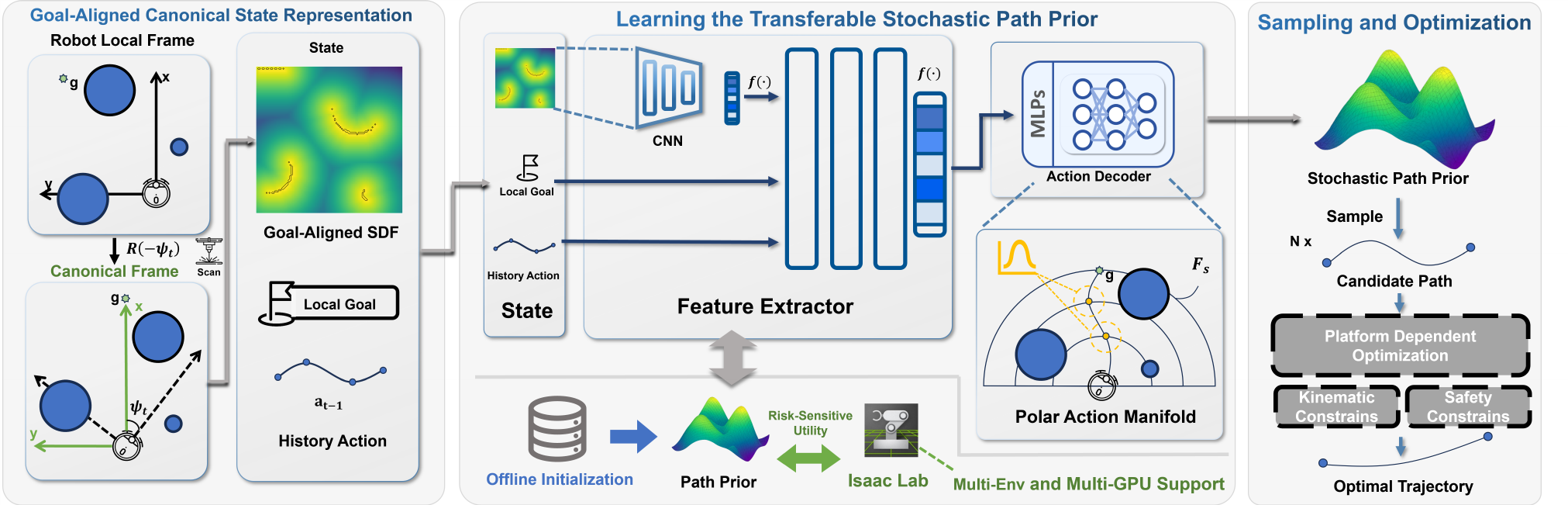}
\vspace{-1.5em}
\caption{
Overview of \method. The left module is a goal-aligned canonical state representation. A transferable goal-conditioned stochastic path prior is learned on a geometry-aware polar action manifold through offline initialization and online single-step interaction refinement with risk-sensitive utility shaping. During inference, sampled paths are refined with constrained trajectory optimization to produce safe, feasible motion.
}
\label{fig:observation}
\vspace{-1.5em}
\end{figure*}

\section{METHOD}

Fig.~\ref{fig:observation} illustrates the framework of \method. We address real-time navigation by learning a transferable goal-conditioned stochastic path prior that models a reusable distribution of geometry-consistent local paths. To transfer across robot embodiments, 
%the prior focuses on geometric path reasoning from local perception and goal direction, while a separate optimization stage enforces embodiment-specific feasibility and safety constraints. 
we learn the geometric path prior based on a new goal-aligned canonical state representation, which
%We introduce a goal-aligned canonical representation that 
normalizes local geometry by the goal direction. This removes in-plane rotational ambiguity and enables flexible provisional-goal selection at test time while retaining the same prior. To stabilize learning of expressive path distributions, we further parameterize local paths on a geometry-aware polar action manifold with radially ordered constraints and incorporate risk-sensitive utility shaping to promote safety-aware behavior. During inference, we use circle-based multi-goal sampling and distributional rollouts to generate diverse candidate paths, then apply a sample-and-refine procedure to obtain a final executable trajectory under hard constraints.

\subsection{Problem Formulation}

We formulate real-time navigation as learning a transferable goal-conditioned stochastic path prior. Given a goal-aligned local geometric context $s$, the prior defines a distribution over structured path parameters:
\begin{equation}
\pi_\theta(a\mid s)\;=\;\mathcal{N}\!\big(\mu_\theta(s),\,\Sigma_\theta(s)\big),
\end{equation}
where $a$ encodes a local path in a low-dimensional polar control-point manifold. A sampled parameter vector $a$ is deterministically decoded into a geometric path $\tau=\mathcal{D}(a)$.

The prior is trained to maximize the expected utility of decoded paths:
\begin{equation}
\theta^\star
=
\arg\max_\theta\;
\mathrm{E}_{s\sim p(s)}
\,\mathrm{E}_{a\sim \pi_\theta(\cdot\mid s)}
\Big[
U\big(s,\mathcal{D}(a)\big)
\Big],
\end{equation}
where $p(s)$ denotes the distribution of goal-aligned local contexts induced by navigation, and $U(\cdot)$ measures path length, smoothness, and risk sensitivity.

At deployment, a sampled geometric path $\tau$ is first converted into a time-parameterized seed trajectory $\hat{\xi}=\mathcal{S}(\tau),$ where $\mathcal{S}(\cdot)$ denotes the seed construction procedure. The final executable trajectory is then obtained by solving a platform-dependent constraint-aware optimization problem
\begin{equation}
\xi = \mathcal{O}(\hat{\xi};\mathcal{K}),
\end{equation}
where $\mathcal{K}$ denotes the set of platform-dependent constraints and $\mathcal{O}(\cdot;\mathcal{K})$ enforces hard safety and kinematic feasibility. This separation keeps $\pi_\theta$ focused on geometry-consistent, embodiment-agnostic path structure, while platform-specific execution is handled in the refinement stage.

\subsection{Goal-Aligned Canonical State Representation}
To learn a transferable geometric path prior, we propose a novel goal-aligned canonicalization.
%Goal-aligned canonicalization constructs the state representation used by the stochastic path prior. 
At time $t$, the robot observes a 2D LiDAR point cloud $P_t$ and a local goal $g_t$ specified in the robot body frame $\mathcal{F}_r$. Let $g_t=(r_t,\psi_t)$ denote the goal in polar coordinates in $\mathcal{F}_r$, where $r_t$ is the goal range and $\psi_t$ is the goal direction. We define a canonical state frame $\mathcal{F}_s$ whose origin coincides with the robot and whose $x$-axis aligns with the ray from the robot to $g_t$. This goal-aligned rotation maps any point $x^{(r)}\in\mathcal{F}_r$ into $\mathcal{F}_s$ as
\begin{equation}
x^{(s)} \;=\; R(-\psi_t)\,x^{(r)}, \qquad R(\cdot)\in SO(2),
\end{equation}
under which the goal becomes $g_t^{(s)}=(r_t,0)$. Reconstructing $\mathcal{F}_s$ at each step removes in-plane rotational ambiguity and normalizes local geometry by the current goal direction. % \looseness=-2

% \begingroup
% \looseness=-1
We encode local geometry using a single-frame signed distance field (SDF) $D_t$ defined on a fixed $H\times W$ grid centered at the robot in $\mathcal{F}_s$. To compute $D_t$, we rotate the LiDAR cloud $P_t$ from $\mathcal{F}_r$ to $\mathcal{F}_s$ using $R(-\psi_t)$, crop it to a fixed window aligned with the $x$-axis of $\mathcal{F}_s$, rasterize it into an occupancy grid, and apply a Euclidean signed distance transform. Each grid cell stores the signed distance to the nearest obstacle, positive in free space and negative inside obstacles, with optional clipping and normalization. This goal-aligned SDF captures the obstacle structure most relevant to local progress. % \looseness=-3
% \par
% \endgroup

The goal condition enters the state through the range-only term $\rho_t=r_t$. Because the goal direction is always zero in $\mathcal{F}_s$, no angular component is needed, yielding a compact representation invariant to in-plane rotations of $\mathcal{F}_r$. We also include the previous action $a_{t-1}$ to promote temporal consistency. The resulting canonical state is
% \vspace{-0.1em}
\begin{equation}
s_t \;=\; \big[\, D_t,\; \rho_t,\; a_{t-1}\,\big],
\end{equation}
%\vspace{-0.1em}
where all quantities are expressed in $\mathcal{F}_s$. 
This canonicalization is important for transfer and reuse. Different provisional goals correspond to different directions $\psi_t$ in $\mathcal{F}_r$, yet all are rotated into the same goal-aligned frame $\mathcal{F}_s$ before being passed to the prior. As a result, the same goal-conditioned path prior $\pi_\theta(a\mid s)$ can be queried repeatedly under circle-based multi-goal sampling by re-normalizing the observation into $\mathcal{F}_s$ for each provisional goal, while platform-dependent feasibility is handled separately during optimization. 

\subsection{Geometry-Aware Polar Action Manifold}

We represent a local path by $n$ polar control points defined in the goal-aligned canonical frame $\mathcal{F}_s$:
\begin{equation}
a=\big[(r_1,\theta_1),\ldots,(r_n,\theta_n)\big]\in \mathrm{R}^{2n},
\end{equation}
These control points are decoded into a continuous geometric trajectory $\tau=\mathcal{D}(a)$ through cubic spline interpolation. To encode geometric structure directly in the action space, we define a geometry-aware polar action manifold by imposing radially ordered constraints:
\begin{equation}
r_i^{\max}=\frac{i}{n+1}\,r_{\max},\quad i=1,\ldots,n,
\end{equation}
together with an angular feasibility cone $\theta_{\min}\le\theta_i\le\theta_{\max}$. These constraints define a structured, ordered subset of $\mathrm{R}^{2n}$ that captures geometry-consistent path hypotheses.

Instead of predicting deterministic paths, we learn a transferable goal-conditioned stochastic path prior over this structured action space. The policy is defined over a normalized latent variable $z\in\mathrm{R}^{2n}$ and modeled as a Gaussian. Following standard continuous-control practice, Gaussian samples are squashed through a $\tanh$ transformation to produce bounded normalized actions $a^{\text{norm}}=\tanh(z)\in[-1,1]^{2n}.$
The normalized actions are then deterministically mapped onto the geometry-aware polar manifold as
\begin{equation}
r_i = \frac{r_i^{\max}}{2}(a^{\text{norm}}_{r_i}+1),
~~
\theta_i = \theta_{\min} + \frac{\theta_{\max}-\theta_{\min}}{2}(a^{\text{norm}}_{\theta_i}+1).
\end{equation}

This construction injects geometric priors into the stochastic policy while preserving differentiability and Gaussian exploration. As a result, the learned prior operates on a reduced structured action manifold, supporting geometry-consistent sampling during inference.

\subsection{Learning the Transferable Stochastic Path Prior}
\label{sec:prior_learning}

We train the goal-conditioned stochastic path prior $\pi_\theta(a\mid s)$ in two stages. In the first stage, we fit a stochastic geometric proposal distribution from large-scale synthesized expert paths. In the second stage, we refine sampling preference through interaction using a trajectory-level risk-sensitive utility.

\subsubsection{Stage I: Offline maximum-likelihood initialization}
\label{sec:prior_stage1}

%We synthesize collision-free, geometry-consistent paths with physics-informed planners ~\cite{ntfields,pntfields,pcplanner} \zzy{for physically-constrained optimal planning} and efficient batch trajectory generation.
%efficient batch trajectory generation. 
% \zp{Why physics}
In this paper, we adopt recent physics-informed planners~\cite{ntfields,pntfields,pcplanner} to synthesize collision-free, geometry-consistent expert paths, leveraging their ability to perform physically constrained optimal planning and efficient batch trajectory generation.
%We choose PC-Planner~\cite{pcplanner} for richer geometric representation, the pipeline generates over $10^6$ training trajectories within 30 minutes without human demonstrations. 
Among them, we use PC-Planner~\cite{pcplanner} due to its richer geometric representation. 
%It can generate more than $10^6$ training trajectories within 30 minutes without human demonstrations.
Using this planner, more than $10^6$ training trajectories are generated within 30 minutes without human demonstrations.
Each expert path is converted into a training pair $(s,a)$, where $s$ is the goal-aligned canonical state and $a\in\mathrm{R}^{2n}$ is the polar control-point parameter defined on the structured manifold. We initialize the Gaussian prior $\pi_\theta(a\mid s)$ by maximum likelihood using the latent actions associated with feasible paths generated in Stage~I. The training objective consists of three terms: the negative log-likelihood, an entropy regularization term, and an $\ell_2$ regularization term. During this stage, the previous latent action is set to $a_{t-1}=0$. % \looseness=-5

\subsubsection{Stage II: Interaction driven refinement with risk sensitive utility}
\label{sec:prior_stage2}

In this stage, we further improve the transferable goal-conditioned stochastic path prior using one-step execution feedback while keeping learning in the space of path distributions. Given a context $s$, an action is sampled as $a\sim\pi_\theta(\cdot\mid s)$ and decoded into a geometric path $\tau=\mathcal{D}(a)$. Each trial produces one executed rollout, reducing learning to a contextual bandit update rather than a multi-step MDP. Across trials, the robot is reset to a fixed start state while the nominal local goal advances along a global reference route, producing controlled coverage of nearby goal-conditioned contexts. 
%Each trial is evaluated using a utility defined over the executed spatial path $\tau$, \zp{which considers complementary aspects of navigation behavior, including safety, efficiency, geometric smoothness, and temporal stability.}
Each trial is evaluated using a utility defined over the executed spatial path $\tau$, which captures complementary aspects of navigation behavior, including safety, efficiency, geometric smoothness, and temporal stability, as summarized in Table~\ref{tab:stage2_reward}. 
%Although execution is time-parameterized, all shaping terms are computed over the spatial footprint of the path. 
%These terms regularize complementary aspects of navigation behavior, including safety, efficiency, geometric smoothness, and temporal stability.
All shaping utility terms are normalized through the squashing function
\begin{equation}
\phi(x;\alpha,\beta)=\tanh\!\Big(\beta\,\frac{x}{\alpha+\varepsilon}\Big),
\qquad x\ge0,
\label{eq:phi_short}
\end{equation}
where $\varepsilon=10^{-6}$. 
%Each utility term is summarized together with the associated expressions in Table~\ref{tab:stage2_reward}. 
The resulting normalized utilities are aggregated into a weighted utility function that evaluates each rollout. Given the success indicator $u_{\mathrm{succ}}$ and the set of shaping utilities $\{u_j\}_{j\in\mathcal{J}}$, the overall utility is
\begin{equation}
U(s,\tau)=
u_{\mathrm{succ}}+\sum_{j\in\mathcal{J}}\lambda_j\,u_j,
\label{eq:utility_sum_short}
\end{equation}
% \vspace{-0.5em}
where $\mathcal{J}=\{\mathrm{safe},\mathrm{len},\mathrm{curv\_var},\mathrm{len\_var},\mathrm{temp}\}$ and $\lambda_j$ denotes the weight associated with each term.

Although the robot executes a time-parameterized trajectory, all shaping terms are computed using the spatial footprint of the path. Let $\{q_i\}_{i=1}^{N}$ denote points sampled along $\tau$ in the canonical frame, and let $D_t(\cdot)$ denote the goal-aligned signed distance field evaluated on the inflated map with sampling spacing equal to $0.04$ times the map resolution. The safety term evaluates clearance violations using the spatial margin
\begin{equation}
m(q_i)=\max\!\bigl(0,\,c_{\mathrm{clr}}-D_t(q_i)\bigr),
\label{eq:margin_short}
\end{equation}
where $c_{\mathrm{clr}}=0.6\,\mathrm{m}$. Risk sensitivity is introduced through $\mathrm{CVaR}_\eta$~\cite{conditional} with $\eta=0.10$, which emphasizes the worst clearance violations along the path.

\begin{table}[!t]
\caption{Utility terms used in interaction driven refinement.}
\vspace{-0.5em}
\label{tab:stage2_reward}
\centering
\small
\setlength{\tabcolsep}{4pt}
\renewcommand{\arraystretch}{0.95}

\begin{tabularx}{\columnwidth}{>{\footnotesize}l >{\footnotesize$}X<{$}}
\toprule
Utility term & \text{Equation} \\
\midrule

Success signal
& u_{\mathrm{succ}}\in\{0,1\} \\

Safety
& u_{\mathrm{safe}}(\tau)=-\phi(\mathrm{CVaR}_\eta[m_i]) \\

Path length
& u_{\mathrm{len}}(\tau)=-\phi(\delta_L) \\

Curvature variation
& u_{\mathrm{curv\_var}}(\tau)=-\phi(\delta_\kappa^{\mathrm{var}}) \\

Segment length variation
& u_{\mathrm{len\_var}}(\tau)=-\phi(\delta_\ell^{\mathrm{var}}) \\

Latent action change
& u_{\mathrm{temp}}(a_t,a_{t-1})=-\phi(\|a_t-a_{t-1}\|_2) \\

\bottomrule
\end{tabularx}
\vspace{-2.0em}
\end{table}

The path length term uses the normalized excess length
\begin{equation}
\delta_L=
\frac{\max(L-\rho_t,0)}{\max(\rho_t,\rho_{\min})+\varepsilon},
\end{equation}
where $\rho_{\min}=0.8\,\mathrm{m}$. The squashing parameters are $(\alpha,\beta)=(0.15,0.74)$ with weight $\lambda_{\mathrm{len}}=1.59$. The curvature variation statistic is
\begin{equation}
\delta_{\kappa}^{\mathrm{var}}=
\frac{1}{N-3}\sum_i
\bigl|\kappa_{i+1}-\kappa_i\bigr|,
\end{equation}
with squashing parameters $(\alpha,\beta)=(0.36,0.56)$ and weight $\lambda_{\mathrm{curv\_var}}=0.98$. The segment-length variation statistic is
\begin{equation}
\delta_{\ell}^{\mathrm{var}}=
\frac{\frac{1}{N-1}\sum_i |d_{i+1}-d_i|}{\bar d+\varepsilon},
\end{equation}
and the squashing parameters are $(\alpha,\beta)=(0.0409,0.96)$ with weight $\lambda_{\mathrm{len\_var}}=1.34$. The latent action change term uses $\|a_t-a_{t-1}\|_2$ with squashing parameters $(\alpha,\beta)=(0.32,0.82)$ and weight $\lambda_{\mathrm{temp}}=1.78$. % \looseness=-1

The policy $\pi_\theta$ is refined using PPO~\cite{ppo} to maximize the expected utility. Under the single-step interaction setting, each trial yields a sample $(s,a)$ with an associated utility, and the PPO objective reduces to
% \vspace{-1.2em}
\begin{equation}
\max_\theta
\mathrm{E}
\left[
\min(\rho U,\bar\rho U)
\right],
\label{eq:ppo_degenerate}
\end{equation}
% \vspace{-0.9em}
\begin{equation}
\rho=
\frac{\pi_\theta(a|s)}
{\pi_{\theta_{\mathrm{old}}}(a|s)},
~~
\bar\rho=\mathrm{clip}(\rho,1-\epsilon,1+\epsilon).
\label{eq:ppo_ratio_clip}
\end{equation}
% \vspace{-0.25 em}
This objective keeps updates close to the data-collecting prior $\pi_{\theta_{\mathrm{old}}}$ while increasing the probability mass assigned to latent samples whose decoded paths yield higher execution utility. % \looseness=-1

\begin{table}[!t]
    \caption{Performance of navigation methods across environments (\textbf{Env.}); \textbf{Meth.} denotes the method. \textbf{Bold} and \underline{underline} mark the best and second best, respectively. \emph{Average} is shown only when all environments have valid results, otherwise ``--''.}

    \renewcommand{\arraystretch}{0.95}
    \vspace{-0.8em}
    \centering
    \resizebox{\columnwidth}{!}{%
    \begin{tabular}{clcccc}
    \toprule
    \textbf{Env.} & \textbf{Meth.} & \textbf{SR}~(\(\uparrow\)) & \textbf{Len. (m)}~(\(\downarrow\)) & \textbf{Time (s)}~(\(\downarrow\)) & \textbf{Vel. (m/s)}~(\(\uparrow\)) \\
    \midrule
    \multirow{4}{*}{\shortstack{Autolab\\35~pedestrians}}
    & DRL-VO      & 0.83 & 5.93 & \textbf{12.86} & \underline{0.46} \\
    & PathRL      & 0.79 & 7.76 & 16.23 & \textbf{0.48} \\
    & CrowdSurfer & \underline{0.86} & \underline{5.69} & \underline{13.27} & 0.43 \\
    & AutoPath        & \textbf{0.94} & \textbf{5.60} & 13.96 & 0.40 \\
    \midrule
    \multirow{4}{*}{\shortstack{Autolab\\55~pedestrians}}
    & DRL-VO      & \underline{0.80} & 6.09 & \textbf{13.43} & \underline{0.45} \\
    & PathRL      & 0.68 & 7.87 & 16.56 & \textbf{0.47} \\
    & CrowdSurfer & \underline{0.80} & \underline{5.84} & \underline{14.34} & 0.41 \\
    & AutoPath        & \textbf{0.95} & \textbf{5.75} & 15.10 & 0.38 \\
    \midrule
    \multirow{4}{*}{\shortstack{Cumberland\\35~pedestrians}}
    & DRL-VO      & 0.82 & 5.83 & \underline{12.99} & \textbf{0.45} \\
    & PathRL      & 0.77 & 6.30 & 15.00 & \underline{0.43} \\
    & CrowdSurfer & \underline{0.85} & \textbf{5.53} & \textbf{12.86} & \underline{0.43} \\
    & AutoPath        & \textbf{0.94} & \underline{5.68} & 13.70 & 0.42 \\
    \midrule
    \multirow{4}{*}{\shortstack{Cumberland\\55~pedestrians}}
    & DRL-VO      & 0.74 & 6.01 & \underline{14.13} & \underline{0.43} \\
    & PathRL      & 0.76 & 6.50 & \textbf{13.74} & \textbf{0.47} \\
    & CrowdSurfer & \underline{0.78} & \textbf{5.65} & 14.29 & 0.40 \\
    & AutoPath        & \textbf{0.93} & \underline{5.93} & 14.99 & 0.40 \\
    \midrule
    \multirow{4}{*}{\shortstack{Freiburg\\35~pedestrians}}
    & DRL-VO      & 0.78 & 6.33 & 15.39 & \underline{0.42} \\
    & PathRL      & 0.71 & 6.94 & \textbf{14.73} & \textbf{0.47} \\
    & CrowdSurfer & \underline{0.80} & \textbf{5.92} & 15.28 & 0.39 \\
    & AutoPath        & \textbf{0.93} & \underline{5.96} & \underline{14.89} & 0.40 \\
    \midrule
    \multirow{4}{*}{\shortstack{Freiburg\\55~pedestrians}}
    & DRL-VO      & \underline{0.72} & 6.13 & \textbf{13.84} & \underline{0.44} \\
    & PathRL      & 0.54 & 7.95 & 17.07 & \textbf{0.46} \\
    & CrowdSurfer & 0.68 & \textbf{5.93} & 17.84 & 0.33 \\
    & AutoPath        & \textbf{0.88} & \underline{6.12} & \underline{16.42} & 0.37 \\
    \midrule
    \multirow{4}{*}{\shortstack{Random\\dense static}}
    & DRL-VO             & --   & --   & --    & --    \\
    & PathRL             & --   & --   & --    & --    \\
    & CrowdSurfer        & \underline{0.38} & \underline{9.83} & \underline{22.03} & \textbf{0.45} \\
    & AutoPath               & \textbf{1.0} & \textbf{7.87} & \textbf{18.22} & \underline{0.43} \\
    \midrule
    \multirow{4}{*}{\shortstack{Average}}
    & DRL-VO      & -- & --   & --    & --    \\
    & PathRL      & -- & --   & --    & --    \\
    & CrowdSurfer & \underline{0.74} & \underline{6.34} & \underline{15.70} & \textbf{0.41} \\
    & AutoPath        & \textbf{0.94} & \textbf{6.13} & \textbf{15.33} & \underline{0.40} \\
    \bottomrule
    \end{tabular}}
    \vspace{-2.0 em}
    \label{tab:all_envs}
\end{table}

{\sloppy
\subsection{Multi-Goal Sampling and Platform-Dependent Optimization}
}
\label{sec:sample_refine}
Feasible local motion in cluttered environments is multimodal. \zzy{Consistent with the canonical state design, we sample diverse candidate goals and paths from the learned stochastic prior to explore multiple feasible motion modes.} Our planning has two steps: path generation from the learned stochastic prior through multi-goal sampling to explore alternative smooth, collision-aware and feasible paths, followed by trajectory optimization under platform-specific constraints. The stochastic prior $\pi_\theta(a\mid s)$, path decoder $\mathcal{D}(\cdot)$, and multi-goal sampling procedure are shared across platforms. Platform differences are specified in trajectory optimization through the constraint set $\mathcal{K}$ and a small set of platform-related optimization parameters. % \looseness=-3

At each planning cycle, we construct a circle centered at the robot with radius $\rho_t$, where $\rho_t$ denotes the distance to the nominal local goal encoded in the canonical state. We uniformly sample $K$ points on this circle as provisional local goals $\{g_k\}_{k=1}^{K}$. For each $g_k$, a goal-aligned canonical frame is instantiated, and the canonical context $s_k$ is constructed by rotating the local observation into $\mathcal{F}_s$.
Conditioned on $s_k$, we draw $M$ samples from the stochastic path prior and decode each sample into a geometric path:
\begin{equation}
a_k^{(m)} \sim \pi_\theta(\cdot\mid s_k), \qquad
\tau_k^{(m)}=\mathcal{D}\!\big(a_k^{(m)}\big),
\label{eq:sample_decode}
\end{equation}
where $\tau_k^{(m)}$ denotes a geometric path in the workspace. The resulting candidate set is
\begin{equation}
\mathcal{C}
=
\bigcup_{k=1}^{K}\,\big\{\tau_k^{(m)}\big\}_{m=1}^{M},
\label{eq:candidate_set}
\end{equation}
where $k$ indexes the sampled provisional goals and $m$ indexes the sampled paths for each goal.

For each sampled path $\tau_k^{(m)}$, we first construct a time-parameterized seed trajectory. Specifically, the path is discretized into $N$ points $\{p_i\}_{i=1}^{N}$ and re-parameterized by arc length. The points are then mapped onto a fixed time horizon $T$ as  % \looseness=-1
\begin{equation}
t_i
=
T
\frac{\sum_{j=2}^{i}\|p_j-p_{j-1}\|}
{\sum_{j=2}^{N}\|p_j-p_{j-1}\|},
\qquad i=1,\ldots,N,
\label{eq:arc_time}
\end{equation}
where $\{(t_i,p_i)\}$ forms a time-indexed waypoint sequence. A 10th-order Bernstein trajectory is then fitted to these waypoints by least squares, yielding an initial seed trajectory $\hat{\xi}_k^{(m)}$.

% The seed trajectory is optimized under the platform-specific constraint set $\mathcal{K}$:\looseness=-4
% \begin{equation}
% \xi_k^{(m)}=\mathcal{O}\!\left(\hat{\xi}_k^{(m)};\mathcal{K}\right),
% \label{eq:refine_each}
% \end{equation}
% where $\xi_k^{(m)}$ denotes the optimized time-parameterized trajectory and $\mathcal{O}(\cdot;\mathcal{K})$ denotes the trajectory optimization operator. 

When deploying on a different platform, we keep the transferable path prior $\pi_\theta(a\mid s)$, the path decoder $\mathcal{D}(\cdot)$, the multi-goal sampling procedure, and the seed construction map $\hat{\xi}=\mathcal{S}(\tau)$ unchanged, and re-specify only the platform-dependent constraint set $\mathcal{K}$ in the refinement stage $\xi=\mathcal{O}(\hat{\xi};\mathcal{K})$. In practice, $\mathcal{K}$ includes kinematic limits such as maximum velocity and acceleration, nominal speed, temporal parameters such as planning horizon and trajectory length, and footprint-related obstacle inflation parameters defined by the obstacle semi-axes. We instantiate $\mathcal{O}(\cdot;\mathcal{K})$ with Priest~\cite{priest}, \zzy{which takes the Bernstein seed trajectory $\hat{\xi}$ constructed from smooth and collision-aware paths sampled from the prior as input and refines it into an executable trajectory $\xi$} through elite sampling and constraint projection to satisfy obstacle avoidance and kinematic constraints.

\vspace{1.0 em}
\section{Experiments}

In this section, we evaluate the proposed method in simulation and real-world settings. We first test it in Gazebo and Isaac Sim across diverse 3D environments and then deploy it on a real TurtleBot4 and a Go2 quadruped.

% \vspace{-0.5 em}
\subsection{Experimental Setup}

\subsubsection{Simulation Configurations}

\begin{figure}[!tbp]
\centering
    \subfloat[]{
        \includegraphics[width=0.35\linewidth]{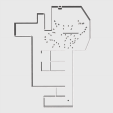}
        \label{fig:cumberland}
     }
     \hspace{0.025\linewidth}
     \subfloat[]{
        \includegraphics[width=0.35\linewidth]{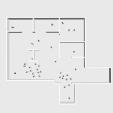}
        \label{fig:autolab}
     }
     \\
     \vspace{-0.8em}
     \subfloat[]{
        \includegraphics[width=0.35\linewidth]{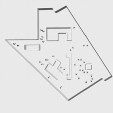}
        \label{fig:freiburg}
     }
     \hspace{0.025\linewidth}
     \subfloat[]{
        \includegraphics[width=0.35\linewidth]{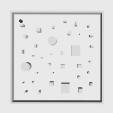}
        \label{fig:random}
     }
     \\
    \caption{
        Gazebo test environments with different scene sizes and pedestrian densities:
        (a) Cumberland, (b) Autolab, (c) Freiburg, and (d) Random.
        Cumberland, Autolab, and Freiburg each measure $20 \times 10$ m and include two pedestrian densities with 35 and 55 agents.
        The Random environment measures $10 \times 10$ m and contains randomly generated layouts.
     }
    \label{fig:anotheu_image}
    \vspace{-1.5em}
\end{figure}

We consider four evaluation scenarios, all unseen during training, as shown in Fig.~\ref{fig:anotheu_image}. Three are public SFM-driven~\cite{sfm} dynamic benchmarks, Autolab, Cumberland, and Freiburg, from \cite{drl_low1}, and the fourth is a randomly generated static scenario. For the dynamic cases, PEDSIM~\cite{pedsim} simulates two crowd densities with 35 and 55 pedestrians. To reduce variance from overly fast motion, we set the mean pedestrian speed to 1.2 m/s and the standard deviation to 0.26 m/s. The experiments use a TurtleBot2 equipped with a ZED stereo camera and a Hokuyo UTM-30LX LiDAR. The maximum TurtleBot2 velocity is 0.5 m/s. The ZED camera provides depth measurements from 0.3 m to 20 m with a $90^\circ$ field of view, and the Hokuyo LiDAR covers 0.1 m to 30 m over $270^\circ$ with an angular resolution of $0.25^\circ$. \zzy{In simulation, each planning cycle uses 20 provisional local goals and 110 candidate paths. The optimizer uses 2 outer and 13 inner iterations with a 4.0~s prediction horizon discretized into 50 points, and the SDF is $105 \times 105$ over $x\in[0,4.2]$~m and $y\in[-2.1,2.1]$~m.} All experiments run on a workstation with an Intel i7-13700KF CPU and an RTX 4070 GPU. % \looseness=-1

\subsubsection{Baselines and Evaluation Metrics}

We compare the proposed approach with three recent learning-based planning methods. Two are DRL-based: DRL-VO~\cite{drl_low1}, which outputs low-level controls, and PathRL~\cite{pathrl}, which generates constrained local paths. Both outperform earlier model-based approaches. The third baseline, CrowdSurfer~\cite{crowdsurfer}, is a hybrid method that uses a generative model to produce prior paths and then optimizes them into the final trajectory. We evaluate navigation with four metrics: SR (success rate), Time (average traversal time in seconds), Len. (average path length in meters), and Vel. (average velocity in m/s). For each scenario, we use the same 25 waypoints as the benchmark~\cite{drl_low1} for the dynamic settings and run four trials under identical initial conditions for a fair comparison. % \looseness=-6

\vspace{-0.5ex}
\subsection{Evaluation on 3D Simulation Scenarios}

\subsubsection{Dynamic Environments}
We first evaluate navigation amid dynamic pedestrian flows in Autolab, Cumberland, and crowded Freiburg. The experiments cover moderate-density settings with 35 pedestrians and high-density settings with 55 pedestrians. In terms of success rate, our policy outperforms all baselines across dynamic scenarios and maintains success rates around $90\%$ as pedestrian density increases. Although demonstrations are collected from one representative training environment, \method generalizes well to all unseen scenes. This generalization stems from the goal-aligned canonical state representation and learned goal-conditioned stochastic path prior, which reduce irrelevant variation and provide reusable proposals for test-time sampling and optimization-based refinement. Compared with
%representative baselines including 
DRL-VO, PathRL, and CrowdSurfer, \method 
improves the success rate by at least 11\%, 15\%, and 8\%, respectively, across all tested environments.
%consistently achieves higher success rates, improving performance \zp{by at least 11\%, 15\%, and 8\%, respectively,}
%by up to $15\%$. 
These gains stem from predicting structured trajectories and leveraging a stochastic path prior with sampling-and-refinement, which improves global consistency, safety, and adaptability under frequent interactions. In Freiburg with 55 pedestrians, our method reaches a success rate of $88\%$, while all baselines remain below $72\%$, highlighting the proposed method's robustness in highly challenging dynamic scenarios. % \looseness=-1

For path length, reflecting navigation efficiency, Table~\ref{tab:all_envs} shows that our method achieves the shortest paths in Autolab while maintaining the highest success rate. In the other dynamic environments, it consistently yields the second-shortest paths, confirming its efficiency. Although our method does not achieve the best time or velocity, its results remain close to those of the fastest approaches across scenarios. This small gap is acceptable because the proposed method prioritizes safety while sacrificing only limited speed.

\subsubsection{High-Density Static Environments}

To further validate the navigation strategy, we also evaluate the method in high-density static environments. As shown in Table~\ref{tab:all_envs}, DRL-VO fails quickly because unstable actions learned solely through reinforcement learning lead to frequent collisions. PathRL also suffers from persistent collisions because its limited action and state spaces restrict performance. % \looseness=-3

CrowdSurfer achieves a success rate of only about $38\%$. By contrast, our goal-aligned canonical state representation together with a flexible action space achieves $100\%$ reliability, while also obtaining near-optimal path length and traversal time and establishing a clear margin over existing state-of-the-art methods. This robustness comes from the proposed sampling-and-refinement pipeline on a goal-aligned canonical state representation. The learned prior provides diverse proposals, and the explicit refinement step turns them into feasible, collision-free solutions while enforcing safety constraints.

\subsection{Ablation Studies}

\subsubsection{Effectiveness of the Goal-Aligned Canonical State Representation}
We first study the proposed goal-aligned canonical state representation. In the ablation setting, the policy is trained and deployed entirely in the robot local frame, without rotating observations into the goal-aligned canonical frame. We directly execute the mean of the predicted prior as the local path without trajectory optimization. As shown in Table~\ref{tab:ablation_representation}, where GACSR denotes the goal-aligned canonical state representation, this variant achieves only a $65\%$ success rate in the random dense static environment, far below that of the full \method framework. The main failure occurs when the goal lies outside the current observation range. Under the local-frame formulation, the policy cannot maintain a consistent target-oriented geometry, and the relation between obstacles and the desired motion direction becomes ambiguous, which substantially degrades performance. By contrast, the goal-aligned canonical state representation removes this ambiguity by aligning the state with the goal direction and thus provides a more stable and transferable representation for planning. % \looseness=-1

\begingroup
\looseness=-4
\subsubsection{Effectiveness of the Prior Learning}
We evaluate the proposed goal-conditioned stochastic path prior. As a baseline, we construct Bern-Samp, which samples candidate paths in the order-10 Bernstein coefficient space using hand-crafted anisotropic Gaussian perturbations, followed by the same trajectory refinement module as \method to produce executable trajectories. The endpoints are fixed at $(0,0)$ and the nominal local goal. We then apply anisotropic Gaussian perturbations with radial-basis correlations to the control-point indices, with larger perturbations in the normal direction and magnitudes scaled by the goal distance. The perturbed coefficients are evaluated with a shared Bernstein basis to generate smooth candidate paths. \zzy{As shown in Table~\ref{tab:ablation_1}, with the same refinement settings, our method's higher success rate shows that the learned goal-conditioned prior provides more reliable path proposals than hand-crafted Bernstein perturbations, while refinement mainly enforces feasibility.}
\par
\endgroup

\begin{table}[!t]
    \caption{Ablation of the goal-aligned canonical state representation in random (dense static) environment.}
    \vspace{-0.8em}
    \centering
    \resizebox{\columnwidth}{!}{%
    \begin{tabular}{
        l
        c
        c
        c
        c
    }
    \toprule
    \renewcommand{\arraystretch}{0.95}
    \textbf{Method} &
      \multicolumn{1}{c}{\textbf{SR}~(\(\uparrow\))} &
      \multicolumn{1}{c}{\textbf{Len.} (m)~(\(\downarrow\))} &
      \multicolumn{1}{c}{\textbf{Time} (s)~(\(\downarrow\))} &
      \multicolumn{1}{c}{\textbf{Vel.} (m/s)~(\(\uparrow\))} \\
    \midrule
    AutoPath w.o. GACSR & 0.65 & 6.78 & 16.60 & 0.41 \\
    AutoPath  & 1.0 & 8.35 & 17.31 & 0.48 \\
    \bottomrule
    \end{tabular}}
    \label{tab:ablation_representation}
    \vspace{-0.8em}
\end{table}

\begin{table}[!t]
    \caption{Performance in Cumberland environment (\textbf{Ped.} denotes the number of pedestrians). \textbf{Bold} indicates the best result. }
    \vspace{-0.8em}
    \centering
    \renewcommand{\arraystretch}{0.9}
    \resizebox{\columnwidth}{!}{%
    \begin{tabular}{clcccc}
    \toprule
    \textbf{Ped.} & \textbf{Method} &
      \multicolumn{1}{c}{\textbf{SR}~(\(\uparrow\))} &
      \multicolumn{1}{c}{\textbf{Len.} (m)~(\(\downarrow\))} &
      \multicolumn{1}{c}{\textbf{Time} (s)~(\(\downarrow\))} &
      \multicolumn{1}{c}{\textbf{Vel.} (m/s)~(\(\uparrow\))} \\
    \midrule
    \multirow{3}{*}{35}
      & Bern-Samp           & 0.87 & 5.73 & \textbf{11.99} & \textbf{0.47} \\
      & AutoPath(NG only)  & 0.90 & 5.74 & 13.45          & 0.43 \\
      & AutoPath             & \textbf{0.94} & \textbf{5.68} & 13.70 & 0.42 \\
    \midrule
    \multirow{3}{*}{55}
      & Bern-Samp           & 0.85 & \textbf{5.78} & \textbf{12.25} & \textbf{0.47} \\
      & AutoPath(NG only)  & 0.89 & 5.91          & 14.90          & 0.40 \\
      & AutoPath             & \textbf{0.93} & 5.93 & 14.99 & 0.40 \\
    \bottomrule
    \end{tabular}}
    \label{tab:ablation_1}
    \vspace{-0.8em}
\end{table}

\begin{table}[!t]
    \caption{Performance in random (dense static) environment. \textbf{Bold} and 
    \underline{underline} indicate the best and second best.}
    \vspace{-0.8em}
    \centering
    \renewcommand{\arraystretch}{0.9}
    \resizebox{\columnwidth}{!}{%
    \begin{tabular}{
        l
        c
        c
        c
        c
    }
    \toprule
    \textbf{Method} &
      \textbf{SR}~(\(\uparrow\)) &
      \textbf{Len. (m)}~(\(\downarrow\)) &
      \textbf{Time (s)}~(\(\downarrow\)) &
      \textbf{Vel. (m/s)}~(\(\uparrow\)) \\
    \midrule
    w.o. $u_{\text{safe}}$    
        & 0.89  
        & 8.60  
        & 18.12 
        & \underline{0.47} \\

    w.o. $u_{\text{len\_var}}$  
        & 0.85  
        & \textbf{7.94}  
        & \textbf{16.51} 
        & \underline{0.47} \\

    w.o. $u_{\text{temp}}$      
        & \underline{0.91}  
        & 9.13  
        & 20.43 
        & 0.45 \\
    Stage I  & 0.82 & \textbf{7.94} & 18.92 & 0.42 \\

    Stage I and II             
        & \textbf{1.00} 
        & \underline{8.35}  
        & \underline{17.31} 
        & \textbf{0.48} \\
    \bottomrule
    \end{tabular}}
    \label{tab:reward_ablation}
    \vspace{-0.8em}
\end{table}

\begin{table}[!t]
\caption{Effect of curvature-variance reward on path smoothness in random (dense static) environment.}
\vspace{-0.8em}
\centering
\small
\setlength{\tabcolsep}{4pt}
\renewcommand{\arraystretch}{0.9}
\begin{tabular}{lc}
\toprule
\textbf{Method} &
\textbf{Turn (deg/m)}~(\(\downarrow\)) \\
\midrule
AutoPath w.o. $u_{\text{curv\_var}}$
    & 168.64 \\
AutoPath
    & 53.32 \\
\bottomrule
\end{tabular}
\label{tab:curv_var_turn}
\vspace{-1.5em}
\end{table}

\subsubsection{Effectiveness of the Two-Stage Learning Framework and Utility Design}
We further evaluate the interaction-based refinement in the training pipeline. To compare a policy trained only with Stage I offline MLE initialization against the full two-stage framework, which combines Stage I offline MLE initialization with Stage II interaction-based refinement, we directly execute the mean of the predicted prior as the local path without the trajectory refinement module. As shown in Table~\ref{tab:reward_ablation}, the policy trained with the full two-stage framework significantly outperforms the policy trained only with Stage I across all metrics, indicating that Stage II improves performance by reshaping the sampling preference of the prior. % \looseness=-1

% \begingroup
% \looseness=1
To understand this improvement, we analyze the contributions of the utility terms through the ablations summarized in Table~\ref{tab:reward_ablation} and Table~\ref{tab:curv_var_turn}. We again directly execute the mean of the predicted prior as the local path without trajectory optimization. Removing $u_{\text{safe}}$ makes the robot approach obstacles more aggressively and perform abrupt avoidance maneuvers. Removing $u_{\text{len\_var}}$ leads to irregular spacing between path points and weakens geometric consistency. Without $u_{\text{temp}}$, the policy exhibits oscillatory behavior because it frequently switches between feasible directions in multimodal situations. Removing $u_{\text{curv\_var}}$ greatly increases the accumulated turning angle per unit distance, causing excessive high-frequency directional changes and degraded smoothness. These results suggest that the gains from interaction-based refinement are closely tied to the structured utility design, because each term regularizes a different aspect of navigation behavior, and together they promote stable and consistent trajectories. % \looseness=-1
% \par
% \endgroup

\begin{figure}[!t]
\centering
\setlength{\abovecaptionskip}{2pt}
\includegraphics[width=\linewidth]{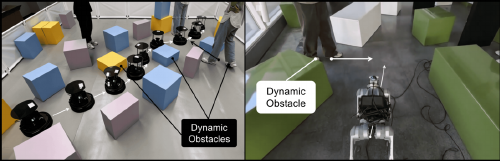}
\caption{
TurtleBot4 and Go2 navigating in real-world environments. Both robots traverse a narrow passage while encountering dynamic obstacles, which makes it harder to reach a target point seven meters away.
}
\label{fig:reality}
\vspace{-1.5 em}
\end{figure}

\subsubsection{Effectiveness of the Circle-Based Multi-Goal Sampling}
We also compare the learned goal-conditioned stochastic path prior with a setting that uses only the nominal local goal, denoted as \method (NG only), to evaluate both the advantage of the prior and the flexibility of multi-goal sampling. As shown in Table~\ref{tab:ablation_1}, circle-based multi-goal sampling consistently yields higher success rates without sacrificing efficiency, demonstrating its benefit.

\subsection{Real-World Deployment and Cross-Platform Transfer}

We deploy \method on a TurtleBot4 Standard operated in safe mode with a maximum linear velocity of 0.31 m/s, and evaluate it in dense static and dynamic scenes. The platform is built on the iRobot Create3 chassis and integrates a 360$^\circ$ RPLIDAR-A1 and an OAK-D Pro depth camera. To evaluate cross-platform transferability, we further deploy the same learned prior without retraining on a real Unitree Go2 quadruped equipped with a Livox Mid-360 LiDAR. As shown in Fig.~\ref{fig:reality}, TurtleBot4 and Go2 successfully navigate complex layouts while avoiding obstacles. \zzy{In addition, we evaluate the policy on ANYmal-D in simulation using the Random environment enlarged twofold, where it achieves a success rate of $96\%$.} Demonstrations of all the above deployments are provided in the supplementary video.
For real-world deployment, to reduce computation, we use one provisional local goal and 50 candidate paths, while keeping other parameters unchanged. On a laptop with a Ryzen 7 7840H CPU and an RTX 4060 Laptop GPU, the local planner runs at approximately 20 Hz. \zzy{In one recorded reference run, the complete local planning pipeline takes 48.25 ms, including 23.37 ms for SDF construction, 0.17 ms for multi-goal sampling, 3.42 ms for prior inference, 11.87 ms for optimization input preparation, 4.28 ms for optimization, and 5.14 ms for other operations.}

\section{Conclusion}

In this paper, we formulate navigation as learning a transferable goal-conditioned stochastic path prior through \method, which models a reusable distribution over geometry-consistent local paths conditioned on local perception and goal direction. A goal-aligned canonical state representation normalizes local geometry with respect to the goal and removes rotational ambiguity, improving training stability, cross-platform transferability, and flexible provisional goal selection. Furthermore, a structured multi-goal sampling strategy combined with distributional path rollouts and trajectory optimization enables rapid generation of diverse and safe paths in dynamic scenarios. Extensive experiments demonstrate that the proposed method achieves high success rates and empirically transfers a single learned path prior \zzy{across the tested differential-drive and quadruped platforms.} Currently, the proposed \method does not explicitly account for interactions between multiple agents. As future work, we plan to extend our approach to handle multi-agent interactions more effectively, thereby improving coordination and efficiency in complex environments.

\bibliographystyle{IEEEtran}
\bibliography{references}

@article{drl_low1,
  title={Drl-vo: Learning to navigate through crowded dynamic scenes using velocity obstacles},
  author={Xie, Zhanteng and Dames, Philip},
  journal={IEEE Transactions on Robotics},
  volume={39},
  number={4},
  pages={2700--2719},
  year={2023},
  publisher={IEEE}
}

@inproceedings{drl_low2,
  title={Intention aware robot crowd navigation with attention-based interaction graph},
  author={Liu, Shuijing and Chang, Peixin and Huang, Zhe and Chakraborty, Neeloy and Hong, Kaiwen and Liang, Weihang and McPherson, D Livingston and Geng, Junyi and Driggs-Campbell, Katherine},
  booktitle={2023 IEEE International Conference on Robotics and Automation (ICRA)},
  pages={12015--12021},
  year={2023},
  organization={IEEE}
}

@inproceedings{drl_low3,
  title={Ldp: A local diffusion planner for efficient robot navigation and collision avoidance},
  author={Yu, Wenhao and Peng, Jie and Yang, Huanyu and Zhang, Junrui and Duan, Yifan and Ji, Jianmin and Zhang, Yanyong},
  booktitle={2024 IEEE/RSJ International Conference on Intelligent Robots and Systems (IROS)},
  pages={5466--5472},
  year={2024},
  organization={IEEE}
}

@article{drl_low4,
  title={Efficient reinforcement learning for autonomous driving with parameterized skills and priors},
  author={Wang, Letian and Liu, Jie and Shao, Hao and Wang, Wenshuo and Chen, Ruobing and Liu, Yu and Waslander, Steven L},
  journal={arXiv preprint arXiv:2305.04412},
  year={2023}
}

@article{trajectory,
  title={Trajectory planning with deep reinforcement learning in high-level action spaces},
  author={Williams, Kyle R and Schlossman, Rachel and Whitten, Daniel and Ingram, Joe and Musuvathy, Srideep and Pagan, James and Williams, Kyle A and Green, Sam and Patel, Anirudh and Mazumdar, Anirban and others},
  journal={IEEE Transactions on Aerospace and Electronic Systems},
  volume={59},
  number={3},
  pages={2513--2529},
  year={2022},
  publisher={IEEE}
}

@inproceedings{pathrl,
  title={Pathrl: An end-to-end path generation method for collision avoidance via deep reinforcement learning},
  author={Yu, Wenhao and Peng, Jie and Qiu, Quecheng and Wang, Hanyu and Zhang, Lu and Ji, Jianmin},
  booktitle={2024 IEEE International Conference on Robotics and Automation (ICRA)},
  pages={9278--9284},
  year={2024},
  organization={IEEE}
}

@inproceedings{pcplanner,
  title={PC-Planner: Physics-Constrained Self-Supervised Learning for Robust Neural Motion Planning with Shape-Aware Distance Function},
  author={Shen, Xujie and Peng, Haocheng and Yang, Zesong and Xu, Juzhan and Bao, Hujun and Hu, Ruizhen and Cui, Zhaopeng},
  booktitle={SIGGRAPH Asia 2024 Conference Papers},
  pages={1--11},
  year={2024}
}

@article{ppo,
  title={Proximal policy optimization algorithms},
  author={Schulman, John and Wolski, Filip and Dhariwal, Prafulla and Radford, Alec and Klimov, Oleg},
  journal={arXiv preprint arXiv:1707.06347},
  year={2017}
}

@article{sfm,
  title={Social force model for pedestrian dynamics},
  author={Helbing, Dirk and Molnar, Peter},
  journal={Physical review E},
  volume={51},
  number={5},
  pages={4282},
  year={1995},
  publisher={APS}
}

@inproceedings{classical_Dlite,
  title={Path planning on robot based on D* lite algorithm},
  author={Belanov{\'a}, Dorota and Mach, Mari{\'a}n and Sin{\v{c}}{\'a}k, Peter and Yoshida, Kaori},
  booktitle={2018 World Symposium on Digital Intelligence for Systems and Machines (DISA)},
  pages={125--130},
  year={2018},
  organization={IEEE}
}

@inproceedings{classical_JPS,
  title={Online graph pruning for pathfinding on grid maps},
  author={Harabor, Daniel and Grastien, Alban},
  booktitle={Proceedings of the AAAI conference on artificial intelligence},
  volume={25},
  number={1},
  pages={1114--1119},
  year={2011}
}

@article{classical_PRM,
  title={Asynchronous multithreading reinforcement-learning-based path planning and tracking for unmanned underwater vehicle},
  author={He, Zichen and Dong, Lu and Sun, Changyin and Wang, Jiawei},
  journal={IEEE Transactions on Systems, Man, and Cybernetics: Systems},
  volume={52},
  number={5},
  pages={2757--2769},
  year={2021},
  publisher={IEEE}
}

@article{classical_RRT,
  title={Rapidly-exploring random trees: Progress and prospects: Steven m. lavalle, iowa state university, a james j. kuffner, jr., university of tokyo, tokyo, japan},
  author={LaValle, Steven M and Kuffner, James J},
  journal={Algorithmic and computational robotics},
  pages={303--307},
  year={2001},
  publisher={AK Peters/CRC Press}
}

@inproceedings{classical_tra1,
  title={Minimum snap trajectory generation and control for quadrotors},
  author={Mellinger, Daniel and Kumar, Vijay},
  booktitle={2011 IEEE international conference on robotics and automation},
  pages={2520--2525},
  year={2011},
  organization={IEEE}
}

@inproceedings{classical_tra2,
  title={Polynomial trajectory planning for aggressive quadrotor flight in dense indoor environments},
  author={Richter, Charles and Bry, Adam and Roy, Nicholas},
  booktitle={Robotics Research: The 16th International Symposium ISRR},
  pages={649--666},
  year={2016},
  organization={Springer}
}

@inproceedings{classical_tra3,
  title={Online generation of collision-free trajectories for quadrotor flight in unknown cluttered environments},
  author={Chen, Jing and Liu, Tianbo and Shen, Shaojie},
  booktitle={2016 IEEE international conference on robotics and automation (ICRA)},
  pages={1476--1483},
  year={2016},
  organization={IEEE}
}

@inproceedings{classical_APF,
  title={Local path planning of mobile robot based on artificial potential field},
  author={Di, Wang and Caihong, Li and Na, Guo and Yong, Song and Tengteng, Gao and Guoming, Liu},
  booktitle={2020 39th Chinese Control Conference (CCC)},
  pages={3677--3682},
  year={2020},
  organization={IEEE}
}

@inproceedings{classical_DWA,
  title={Dynamic window based approach to mobile robot motion control in the presence of moving obstacles},
  author={Seder, Marija and Petrovic, Ivan},
  booktitle={Proceedings 2007 IEEE International Conference on Robotics and Automation},
  pages={1986--1991},
  year={2007},
  organization={IEEE}
}

@article{IL1,
  title={Learning model predictive controllers with real-time attention for real-world navigation},
  author={Xiao, Xuesu and Zhang, Tingnan and Choromanski, Krzysztof and Lee, Edward and Francis, Anthony and Varley, Jake and Tu, Stephen and Singh, Sumeet and Xu, Peng and Xia, Fei and others},
  journal={arXiv preprint arXiv:2209.10780},
  year={2022}
}

@article{IL2,
  title={How to guide your learner: Imitation learning with active adaptive expert involvement},
  author={Liu, Xu-Hui and Xu, Feng and Zhang, Xinyu and Liu, Tianyuan and Jiang, Shengyi and Chen, Ruifeng and Zhang, Zongzhang and Yu, Yang},
  journal={arXiv preprint arXiv:2303.02073},
  year={2023}
}

@article{IL3,
  title={Imitation learning by reinforcement learning},
  author={Ciosek, Kamil},
  journal={arXiv preprint arXiv:2108.04763},
  year={2021}
}

@inproceedings{IL4,
  title={Learning from all vehicles},
  author={Chen, Dian and Kr{\"a}henb{\"u}hl, Philipp},
  booktitle={Proceedings of the IEEE/CVF Conference on Computer Vision and Pattern Recognition},
  pages={17222--17231},
  year={2022}
}

@article{DRL_cost,
  title={Dd-ppo: Learning near-perfect pointgoal navigators from 2.5 billion frames},
  author={Wijmans, Erik and Kadian, Abhishek and Morcos, Ari and Lee, Stefan and Essa, Irfan and Parikh, Devi and Savva, Manolis and Batra, Dhruv},
  journal={arXiv preprint arXiv:1911.00357},
  year={2019}
}

@article{DRL_path,
  title={Robot navigation with reinforcement learned path generation and fine-tuned motion control},
  author={Zhang, Longyuan and Hou, Ziyue and Wang, Ji and Liu, Ziang and Li, Wei},
  journal={IEEE Robotics and Automation Letters},
  volume={8},
  number={8},
  pages={4489--4496},
  year={2023},
  publisher={IEEE}
}

@inproceedings{crowdsurfer,
  title={CrowdSurfer: Sampling Optimization Augmented with Vector-Quantized Variational AutoEncoder for Dense Crowd Navigation},
  author={Kumar, Naman and Singha, Antareep and Nanwani, Laksh and Potdar, Dhruv and Rastgar, Fatemeh and Idoko, Simon and Singh, Arun Kumar and Krishna, K Madhava and others},
  booktitle={2025 IEEE International Conference on Robotics and Automation (ICRA)},
  pages={16854--16860},
  year={2025},
  organization={IEEE}
}

@inproceedings{ac-mpc,
  title={Actor-Critic Model Predictive Control},
  author={Romero, Angel and Song, Yunlong and Scaramuzza, Davide},
  booktitle={2024 IEEE International Conference on Robotics and Automation (ICRA)},
  pages={14777--14784},
  year={2024},
  organization={IEEE}
}

@article{go-mpc,
  title={Where to go Next: Learning a Subgoal Recommendation Policy for Navigation in Dynamic Environments},
  author={Brito, Bruno and Everett, Michael and How, Jonathan P. and Alonso-Mora, Javier},
  journal={IEEE Robotics and Automation Letters},
  volume={6},
  number={3},
  pages={4616-4623},
  year={2021},
  publisher={IEEE}
}

@article{priest,
  title={PRIEST: Projection guided sampling-based optimization for autonomous navigation},
  author={Rastgar, Fatemeh and Masnavi, Houman and Sharma, Basant and Aabloo, Alvo and Swevers, Jan and Singh, Arun Kumar},
  journal={IEEE Robotics and Automation Letters},
  volume={9},
  number={3},
  pages={2630--2637},
  year={2024},
  publisher={IEEE}
}

@misc{pedsim,
  author       = {C. Gloor},
  title        = {PEDSIM: Pedestrian crowd simulation},
  year         = {2016},
  howpublished = {\url{http://pedsim.silmaril.org}},
  note         = {Accessed: 2025-09-11}
}

@article{conditional,
  title={Conditional value-at-risk for general loss distributions},
  author={Rockafellar, R Tyrrell and Uryasev, Stanislav},
  journal={Journal of banking \& finance},
  volume={26},
  number={7},
  pages={1443--1471},
  year={2002},
  publisher={Elsevier}
}

@article{ntfields,
  title={Ntfields: Neural time fields for physics-informed robot motion planning},
  author={Ni, Ruiqi and Qureshi, Ahmed H},
  journal={arXiv preprint arXiv:2210.00120},
  year={2022}
}

@article{pntfields,
  title={Progressive Learning for Physics-informed Neural Motion Planning},
  author={Ni, Ruiqi and Qureshi, Ahmed H},
  journal={arXiv preprint arXiv:2306.00616},
  year={2023}
}

\end{document}